\RequirePackage{fix-cm}
\documentclass[twocolumn]{svjour3}          
\smartqed  
\usepackage{graphicx}
\usepackage{amssymb}
\usepackage[utf8]{inputenc}
\usepackage{xcolor}
\usepackage{hyperref}
\hypersetup{colorlinks}
\usepackage{newfloat}
\definecolor{bg}{rgb}{0.95,0.95,0.95}
\usepackage[cachedir=.,finalizecache=false,frozencache=true]{minted}
\usepackage{multicol}
\begin{document}

\title{OCAPIS: R package for Ordinal Classification And Preprocessing In Scala
}
\subtitle{}


\author{M. Cristina Heredia-Gómez  \and
        Salvador García \and 
        Pedro Antonio Gutiérrez \and
        Francisco Herrera 
}


\institute{M. Cristina Heredia-Gómez \at
              DaSCI Andalusian Institute of Data Science and Computational Intelligence, University of Granada, Spain
              \email{mrcrstnherediagmez@gmail.com}           
           \and
           Salvador García \at
              DaSCI Andalusian Institute of Data Science and Computational Intelligence, University of Granada, Spain
              \email{salvagl@decsai.ugr.es}
           \and
           Pedro Antonio Gutiérrez \at 
           Department of Computer Science and Numerical Analysis, University of Córdoba, Campus de Rabanales, edificio Albert Einstein 14017, Córdoba, Spain 
           \and
           Francisco Herrera \at
           DaSCI Andalusian Institute of Data Science and Computational Intelligence, University of Granada, Spain
}

\date{Received: date / Accepted: date}

\maketitle

\begin{abstract}
Ordinal Data are those where a natural order exists between the labels. The classification and preprocessing of this type of data is attracting more and more interest in the area of machine learning, due to its presence in many common problems. Traditionally, ordinal classification problems have been approached as nominal problems. However, that implies not taking into account their natural order constraints. In this paper, an innovative R package named \textbf{ocapis} (Ordinal Classification and Preprocessing In Scala) is introduced. Implemented mainly in Scala and available through Github, this library includes four learners and two preprocessing algorithms for ordinal and monotonic data. Main features of the package and examples of installation and use are explained throughout this manuscript.
\end{abstract}

\section{Introduction}
\label{intro}
The development of supervised classification and preprocessing techniques for data with numerical targets is a central topic in machine learning and data science \cite{witten2016data,cherkasskylearning}. Nonetheless, it is now that more attention is being paid to classification and preprocessing of ordinal and monotonic data, given their big presence in everyday problems. For example, there is an increasing amount of data from service rating surveys 
whose target is based on an ordinal scale \textit{[ bad, regular, good, very\ good, excellent] }
and therefore class labels incorporate order information, consequently an instance with class \textit{excellent} has a higher rating than another from \textit{regular} class \cite{gutierrez2016ordinal}. Monotonic data is a special case of ordinal data where monotonicity constraints exist between instances and class labels in such a way that given two instances $x, x'$ where $x \le x' \Rightarrow f(x) \le f(x')$. That is, if an instance is smaller or equal than other instance, then its associated class cannot be greater. Monotonic constraints are present in many real data, such as house prices, since they increase directly with the size of the house and the year of construction and inversely with the distance to the city. They are also very present in finance, where there are some companies that dominate others for all financial indicators\cite{cano2017training,gutierrez2016current}. The main challenges when dealing with this kind of data are, on the one hand, considering the ordering information to build more realistic models, and on the other hand, using different misclassification cost depending on error, as labeling an instance as \textit{good} when its real label is \textit{very\ good} is not the same error as labeling it as \textit{bad}. \\
Although there are numerous scientific publications on ordinal classification (also ordinal regression), there are very few open source libraries for performing ordinal and monotonic classification and preprocessing tasks. \newline
For R we find the very recent \textbf{ordinalNet} package \cite{wurm2017ordinalnet}, which fits ordinal regression models with elastic penalty and supports model families from element-wise link
multinomial ordinal class. Similarly, \textbf{ordinal} \cite{christensen2015ordinal}, also implements ordered regression models, commonly named proportional odds models, which allows specifying a link function from \textit{[ logit, probit, loglog, cloglog, cauchit]}, like \textbf{ordinalNet}. \newline
Since last year, \textbf{monMLP} \cite{cannon2015monmlp} is also available, which offers a multi-layer perceptron neural network where monotonicity constraints can be optionally applied. Although there are other packages related to ordinal data, they offer an isolated task or algorithm, like ordinal data conversion \cite{demirtas2016concurrent,toordinal}, mixture models \cite{iannario2016cub}, penalized ratio models \cite{archer2018glmnetcr,glmpathcr}, multiple ordinal tobit models \cite{wright2015multiple}, clustering \cite{selosse2018ordinalclust} or rule models \cite{hornung2017ordinal}. \newline
For Matlab and Octave we find \textbf{orca} \cite{gutierrez2016ordinal}, a more complete library than those mentioned above, which offers many algorithms for ordinal data classification.  \newline

However, there are three main issues with the software mentioned above. First, both \textbf{ordinalNet} and \textbf{ordinal} essentially just offer highly customizable proportional odds models, without considering other techniques. Something similar happens with \textbf{monMLP} that offers multilayer perceptron models. Second, none of them offer preprocessing techniques for ordinal data. Third, although \textbf{orca} offers many classification techniques, reports and configuration capabilities, it is written in MATLAB/octave (which are
less popular languages nowadays for data analysis), the installation
process implies the compilation of different third-party C packages, and
it does not include preprocessing methods (feature and instance selection). \newline
In this paper an innovative and efficient R package named  \textbf{ocapis} \footnote{https://github.com/CristinaHG/OCAPIS} is presented. It is built mainly in Scala \cite{odersky2008programming}, a pretty young JVM language well known for its scalability, mixed paradigm (object-oriented and functional programming) and mixin-composition constructs for composing classes and traits. It also provides object decomposition by pattern matching and a powerful abstraction for types and values, which have made Scala one of the most used languages in Big Data \cite{maillo2017knn}\cite{galicia2018multi}.\newline

Developing \textbf{ocapis} primarily in Scala along with R has been possible by using the very recent \textbf{rscala} \cite{dahlintegration} package. The proposed package is, to our knowledge, the third R package built in Scala after \textbf{shallot} \cite{dahl2017random} and \textbf{bamboo} \cite{li2014bayesian}, both from \textbf{rscala} creator. \textbf{Ocapis} aims to provide an open source library of classification and preprocessing methods for ordinal data that currently lack an implementation in R, including non-linear ordinal classification techniques and one of the most recent instance selector proposed in the literature.\newline
The rest of the manuscript is arranged as follows. \hyperref[sec:2]{Section~\ref*{sec:2}} presents the software and implemented algorithms. \hyperref[sec:3]{Section~\ref*{sec:3}} shows some illustrative examples of use. \hyperref[sec:4]{Section~\ref*{sec:4}} exposes the experimental framework and results. Finally, \hyperref[sec:5]{Section~\ref*{sec:5}} sets out conclusions.

\section{Software}
\label{sec:2}
The importance of creating specific techniques for data of an ordinal nature is beyond all doubt. Since the problem was first studied in statistics by using a link function to model underlying probabilities \cite{anderson1984regression}, the field of ordinal classification has evolved a great deal in recent years \cite{seah2012transductive,tian2014comparative}. In this new package, four of the best-known techniques for ordinal data classification are implemented, along with two preprocessing algorithms, an ordinal feature selector adapted to deal also with monotonic data and a newly proposed instance selector:
\begin{itemize}
	\item svmop. Support Vector Machine with Ordered Partitions (SVMOP classifier) is an ensemble of weighted support vector machines for ordinal regression proposed in \cite{waegeman2009ensemble}, based on Frank \& Hall binary decomposition method \cite{frank2001simple}.
	\item pom. The Proportional Odd Model for Ordinal Regression (POM) is a member of a family of linear models known as cumulative link models or ordered regression models, proposed by \cite{mccullagh1980regression}. It is based on a link function to model class probabilities. Accepted link functions are \textit{( logit, probit, cloglog, loglog, cauchit ),} where \textit{logit} is usually the standard choice.
	\item kdlor. Kernel Discriminant Learning for Ordinal Regression (KDLOR) is a Kernel version of LDA applicable to non-linear data of an ordinal nature. Proposed by \cite{sun2010kernel}, it minimizes the distance within classes and maximizes the distance between classes, while considering the order information of the different classes.
	\item wknnor. Weighted k-Nearest-Neighbor for ordinal classification (WKNNOR) proposed by \cite{hechenbichler2004weighted} maps neighbors distances to weights according to a kernel function. Accepted kernels are: \textit{rectangular, triangular, epanechnikov, biweight, triweight, cosine, gaussian, inversion}. The algorithm has been adapted to cope with monotonic data, incorporating the monotonicity constraints suggested in \cite{duivesteijn2008nearest}.
	\item fselector. This Feature selector for monotonic classification was originally proposed in \cite{hu2012feature}. The preprocessing algorithm is based on Fuzzy Rank Mutual Information (FRMI) \cite{hu2010information} and the search strategy of min-redundancy and max-relevance (mRMR) is used to select best features. 
	\item iselector. Training Set Selection for Monotonic Ordinal Classification. This new proposal \cite{cano2017training} introduces a triphasic instance selector where first, feature selection is performed, then a collision removal is carried out, and finally an evaluation metrics process is applied.
\end{itemize}

\section{Examples of use}
\label{sec:3}
All classification algorithms are designed to have a fit and an analogous predict method. In the following example an ordinal dataset named \textbf{balace-scale} is loaded. Then an example about how to apply the two preprocessing techniques over the training set is given. Finally, we illustrate how to perform classification and prediction using the SVMOP, POM, KDLOR and WKNNOR classifiers.

\begin{minted}
[frame=lines,
framesep=1mm,
fontsize=\footnotesize,
bgcolor=bg,
linenos,
breaklines=true
]{r}
# Load train and test data
data("train_balance-scale")
data("test_balance-scale")
trainlabels<-dattrain[,ncol(dattrain)] # train labs 
traindata=dattrain[,-ncol(dattrain)] # train data
testdata<-dattest[,-ncol(dattest)] # test labels
testlabels<-dattest[,ncol(dattest)] # test data

# Select the three most important features using k and beta=2
selected<-fselector(traindata,trainlabels,2,2,3)
trainselected<-traindata[,selected]

# Select the most relevant instances with a candidate rate=0.02, collision rate=0.1 and considering maximum 5 neighbors
selected<-iselector(traindata,trainlabels,.02,.1,5)
trainselected<-selected[,-ncol(selected)]
trainlabels<-selected[,ncol(selected)]

# Classifying using SVMOP using weights per instance, cost=0.1 and gamma=0.1
modelstrain<-svmofit(traindata,trainlabels,TRUE,.1,
.1)
predictions<-svmopredict(modelstrain,testdata)
sum(predictions[[2]]==testlabels)/nrow(dattest)
[1] 0.9235669

# Classifying using POM with logistic link function
fit<-pomfit(traindata,trainlabels,"logistic")
predictions<-pompredict(fit,testdata)
projections<-predictions[[1]]
predictedLabels<-predictions[[2]]
sum(predictedLabels==testlabels)/nrow(dattest)
[1] 0.910828

# Classifying using KDLOR with RBF kernel, optimization parameter=10, parameter for H matrix=0.001 and kernel param =1
myfit<-kdlortrain(traindata,trainlabels,"rbf",10,
.001,1)
pred<-kdlorpredict(myfit,traindata,testdata)
sum(pred[[1]]==testlabels)/nrow(dattest)
[1] 0.8343949

# Classifying using WKNNOR considering 5 nearest neighbors, euclidean distance, rectangular kernel to compute weights and without monotonicity constraints
predictions<-wknnor(traindata,trainlabels,testdata,
5,2,"rectangular",FALSE)
sum(predictions==testlabels)/nrow(dattest)
[1] 0.7515924
\end{minted}

In the previous example the first 7 lines read the train and test datasets, separating the class labels from the data. In lines 9-11 a feature selection is performed over the training data, choosing the three most relevant features. Similarly, in lines 13-16 an instance selection is performed over the training set. As it returns a complete dataset with the selected instances, we make it our new training set. Then an example of the use of the four implemented classifiers is given. For each of them we start by fitting the model using the training data. After that, predictions are made using the test data. Finally, model accuracy is computed and shown for each model.

\section{Experimental framework and results}
\label{sec:4}
Experiments have been carried out through a comparison of performance and CPU time consumption between the only software solution mentioned above that implements three out of this four classification techniques, \textbf{orca} \cite{gutierrez2016ordinal}, and \textbf{ocapis}. For performance evaluations, two widely used metrics in the field of ordinal classification have been used, named \textbf{MZE} (Mean Zero-one Error) and \textbf{MAE} (Mean Absolute Error). \newline
The Mean Zero-one Error is the error rate of the classifier:

$$MZE=\frac{1}{N} \sum_{i=1}^{N} \left[ \left[ y_i^* \neq y_i \right] \right]= 1 - Accuracy,$$

where $y_i$, $y_i^*$ are the real and predicted values respectively.
This metric ranges from 0 to 1 and relates to global performance, without considering the order.\newline
The MAE is the average deviation in absolute value of the predicted rank from the true one\cite{gutierrez2016ordinal}:

	$$MAE=\frac{1}{N} \sum_{i=1}^{N} |y_i - y_i^*|,$$
where $(y_i,y_i^*)$ represents each real-prediction pair. MAE values range from 0 to $Q-1$, where $Q$ denotes the number of categories, and it uses an absolute cost. \newline


The datasets used for the experiments are described in 
\hyperref[tab:1]{Table~\ref*{tab:1}}. The parameter configuration used is shown in \hyperref[tab:2]{Table~\ref*{tab:2}}. The parameters has been left by default to illustrate performance. \hyperref[tab:3]{Table~\ref*{tab:3}} shows the performance comparison between \textbf{orca} and \textbf{ocapis}, where check-marks represent cases where \textbf{ocapis} performs better than \textbf{orca}.
In order to illustrate the preprocessing techniques behavior, \hyperref[tab:4]{Table~\ref*{tab:4}} and \hyperref[tab:5]{Table~\ref*{tab:5}} show their performance over the mentioned datasets, where check-marks are used to show cases where the preprocessing has shown to improve the base classification results from \hyperref[tab:3]{Table~\ref*{tab:3}}. Finally, \hyperref[tab:6]{Table~\ref*{tab:6}} shows CPU times for \textbf{orca} and \textbf{ocapis} classifiers, while \hyperref[tab:7]{Table~\ref*{tab:7}} shows CPU time for the two preprocessing algorithms implemented in \textbf{ocapis}.

\begin{table}[!h]
	
	\caption{Ordinal datasets used in experiments}
	\label{tab:1}       
	
	\begin{tabular}{llll}
		\hline\noalign{\smallskip}
		Dataset & Instances & Features & Classes  \\
		\noalign{\smallskip}\hline\noalign{\smallskip}
		balance-scale &	625 & 4 & 3 \\
		winequality-red & 1599 & 11 & 6 \\
		SWD & 1000 & 10 & 4 \\
		contact-lenses & 24 & 6 & 3 \\
		toy & 300 & 2 & 5 \\
		ESL & 488 & 4 & 9 \\
		LEV & 1000 & 4 & 5 \\
		Automobile & 205 & 71 & 6\\
		Pasture & 36 & 25 & 3\\
		Squash-stored & 52 & 51 & 3 \\
		\noalign{\smallskip}\hline
	\end{tabular}
\end{table}

\begin{table}[!h]
	\caption{Parameters configuration used for experiments}
	\label{tab:2}       
	\begin{tabular}{l|lll}
		\hline\noalign{\smallskip}
		Algorithm & Configuration  \\
		\noalign{\smallskip}\hline\noalign{\smallskip}
		SVM \cite{chang2011libsvm} & C =.1,$\gamma$=.1\\
		POM \cite{mccullagh1980regression} & logistic linkfunction\\
		KDLOR \cite{sun2010kernel} & RBF kernel, d=10, u=.001, k=1\\
		WKNNOR \cite{hechenbichler2004weighted} & Rectangular kernel, k=5, distance=1\\
		FSelector \cite{hu2012feature} & k, $\beta$ =2, (half of the characteristics)\\
		ISelector \cite{cano2017training} & candidates=.01,collisions=.02, kEd=5\\
		\noalign{\smallskip}\hline
	\end{tabular}
\end{table}

\begin{table}[!h]
	\caption{Times of preprocessing algorithms} \label{tab:7}
	\begin{tabular}{ l l l l l l l l l l l l }
		\hline\noalign{\smallskip}
		Dataset & Feature Selector & Instance selector  \\
		\noalign{\smallskip}\hline\noalign{\smallskip}	
		balance-scale & 1.0900 & 3.4782 \\
		winequality-red & 38.9118 & 8.5956 \\
		SWD & 7.2385 & 4.5675\\
		contact-lenses & 0.0089 & 1.665\\
		toy & 0.3214 & 2.6208 \\
		ESL & 0.9519 & 2.2521 \\
		LEV & 3.8742 & 4.367 \\
		Automobile & 17.1677 & 1.8520 \\
		Pasture & 0.1679 & 1.7769  \\
		Squash-stored & 1.295 & 1.9367\\
		\noalign{\smallskip}\hline  
	\end{tabular} 
\end{table}

\begin{table}[!h]
	\caption{Time comparison between orca and ocapis (seconds) (orca/ocapis)} \label{tab:6}
	\noindent\resizebox{\linewidth}{!}{
		\small 
		\begin{tabular}{ c c c c c c c c c c c l }
			\hline
			Dataset & SVM & POM & KDLOR & WKNNOR  \\
			\hline	
			balance-scale &3.1671/0.2194$\checkmark$ & 0.0733/0.0034$\checkmark$ & 0.5686/0.0022$\checkmark$ & 0.00142 \\
			winequality-red & 2.3509/2.0927$\checkmark$ & 0.0482/0.1222 & 6.0023/6.9004 & 0.1977 \\
			SWD & 1.2154/0.5304$\checkmark$ & 0.0371/0.07723 & 1.2800/1.4996 & 1.4529 \\
			contact-lenses & 1.8219/1.1578$\checkmark$ & 0.0400/0.0500 & 0.1993/0.0164$\checkmark$ & 0.0313  \\
			toy & 1.3535/0.1433$\checkmark$ & 0.0256/0.0220$\checkmark$ & 0.2432/0.1194$\checkmark$ &  0.0288 \\
			ESL & 0.9492/0.4393$\checkmark$ & 0.0287/0.0570 & 0.3676/0.2532$\checkmark$ & 0.0368 \\
			LEV & 1.2184/0.5660$\checkmark$ & 0.0275/0.0609 &  1.1943/1.4839 & 0.0905 \\
			Automobile & 1.0813/0.1780 $\checkmark$& 0.1183/0.1062$\checkmark$ & 0.2706/0.0728 $\checkmark$& 0.0223 \\
			Pasture & 1.0201/0.0208$\checkmark$ & 0.0421/0.0417 $\checkmark$ & 0.2151/0.0117$\checkmark$ & 0.0053  \\
			Squash-stored & 1.1047/0.0245$\checkmark$ & 0.0866/0.0616$\checkmark$ & 0.2152/0.0153$\checkmark$ & 0.0117 \\
			\hline  
		\end{tabular}
	} 
\end{table}

From \hyperref[tab:3]{Table~\ref*{tab:3}} we may conclude that our implementation performs equal and sometimes better than \textbf{orca} algorithms. Main performance differences can be seen in SVMOP, where a lot of check-marks denotes that the SVMOP implemented in \textbf{ocapis} gets better results than the SVMOP implemented in \textbf{orca}. In spite of both use libsvm-weights \cite{chang2011libsvm} implementation underneath, as it is originally implemented in C, one uses the Matlab wrapper while the other uses the Python wrapper. In KDLOR, we can see that \textbf{ocapis} performs exactly equal and in three cases better than \textbf{orca}. The cause is that while \textbf{orca} uses the QP solver from Matlab, \textbf{ocapis} uses the QP solver from the very new Scala library \textbf{Breeze} \cite{hall2009breeze} still under development. Besides that we can see that for large datasets as \textit{Automobile} with 71 features, a preprocessing step is mandatory to reduce the problem dimensionality, as some algorithms like POM may present problems to converge with such amount of features. In this case \textbf{ocapis} is a more complete software option as it offers two preprocessing algorithms while \textbf{orca} does not include any. From \hyperref[tab:4]{Table~\ref*{tab:4}} and \hyperref[tab:5]{Table~\ref*{tab:5}} we can see that classification techniques can greatly benefit from a previous preprocessing step, especially when dealing with datasets where number of features or instances is large. Lastly, from \hyperref[tab:6]{Table~\ref*{tab:6}} we can point a clear advantage in performing times for \textbf{ocapis} over \textbf{orca} in the algorithms implemented in Scala, which are SVMOP, KDLOR and WKNNOR, this difference is not so large for POM, which is implemented in R. In addition, for all of them, \textbf{ocapis} gets much shorter CPU times than \textbf{orca} when dealing with high-dimensional datasets as \textit{Automobile} (71 features), \textit{Pasture} (25 features) and \textit{Squash-stored} (51 features), due to its Scala implementations applying functional programming and immutability principles.\newline Whereas WKNNOR is not included in ORCA and not tested with these datasets in its original proposal \cite{hechenbichler2004weighted}, this Scala implementation has shown a very good time performance even with high-dimensional datasets. From \hyperref[tab:7]{Table~\ref*{tab:7}} we conclude that even though preprocessing is 

\onecolumn
\begin{table}
	\centering
	\caption{Performance comparison between orca and ocapis (orca/ocapis)}
	\label{tab:3}       
	\noindent\resizebox{\linewidth}{!}{
	\begin{tabular}{c c c c c c c c c c c l}
		\cline{2-9}
		&  \multicolumn{2}{c}{SVM}
		&  \multicolumn{2}{c}{POM} 
		&  \multicolumn{2}{c}{KDLOR}
		&  \multicolumn{2}{c}{WKNNOR}  \\
		\hline\noalign{\smallskip}
		Dataset & MAE & MZE & MAE & MZE & MAE & MZE & MAE & MZE \\
		\noalign{\smallskip}\hline\noalign{\smallskip}
		balance-scale &0.0890/0.0890 & 0.0764/0.0760 & 0.1019/0.1019 & 0.0891/0.0891 & 0.1656/0.1656 & 0.1656/0.1656 & 0.4076 & 0.2484 \\
		winequality-red & 0.5120/0.5050$\checkmark$ &0.4325/0.4300$\checkmark$ & 0.4475/0.4425$\checkmark$ & 0.4100/0.4020$\checkmark$ & 0.5000/0.5100 & 0.4470/0.4600$\checkmark$ & 2.6350 & 0.9950 \\
		SWD & 0.4400/0.4400 & 0.4280/0.4280 & 0.4800/0.4800 &0.4640/0.4640 & 0.5560/0.5080$\checkmark$ & 0.4840/0.4560$\checkmark$ & 1.3240 & 0.8400 \\
		contact-lenses &0.3330/0.3330 & 0.3330/0.3330 & 0.5000/ - & 0.3330/ - & 0.5000/0.5000 & 0.5000/0.5000 & 0.5000 & 0.3333 \\
		toy & 0.4930/0.5860 & 0.4270/0.4800 & 0.8800/0.8800  &  0.6670/0.6670 & 0.1460/0.1460 & 0.1460/0.1460 & 1.9333 & 0.8933 \\
		ESL & 0.3850/0.3770$\checkmark$ & 0.3690/0.3600$\checkmark$ & 0.3610/0.3610 & 0.3270/0.3270 & 0.4180/0.3930$\checkmark$ & 0.4016/0.3524$\checkmark$ & 1.8033  &0.8852 \\
		LEV & 0.4640/0.4360$\checkmark$ & 0.4240/0.400$\checkmark$ &  0.4120/0.4120 & 0.3760/0.3760 & 0.4840/0.4840 & 0.4040/0.4200 &  1.4400 & 0.7840 \\
		Automobile & 2.8269/1.1540$\checkmark$ & 0.9810/0.6920$\checkmark$ & - / - & - / - & 1.0192/1.0192 & 0.7307/0.7307 & 2.8269 & 0.9808 \\
		Pasture & 1/0.6670$\checkmark$ & 0.6670/0.6670 & 0.7780/ - & 0.6670/ - & 0.6670/0.6670 & 0.6670/0.6670 & 1 & 0.6667  \\
		Squash-stored & 0.7690/0.7690 & 0.6150/0.6150 & 0.7692/ - & 0.6923/ - & 0.5385/0.5385 & 0.5385/0.5385 &  0.7692 & 0.6150 \\
		\noalign{\smallskip}\hline
	\end{tabular}
}
\end{table}

\begin{table}[!h]
	\caption{Performance of ocapis Feature selector} \label{tab:4}
	\noindent\resizebox{\linewidth}{!}{
		\begin{tabular}{ c c c c c c c c c c c l }
			\cline{2-9}
			&  \multicolumn{2}{c}{SVM}
			&  \multicolumn{2}{c}{POM} 
			&  \multicolumn{2}{c}{KDLOR}
			&  \multicolumn{2}{c}{WKNNOR}  \\
			\hline
			Dataset & MAE & MZE & MAE & MZE & MAE & MZE & MAE & MZE \\
			\hline	
			balance-scale &	0.5159 & 0.2994 & 0.5159 & 0.2994 & 0.4777  & 0.3376  & 0.8089 & 0.4458  \\
			winequality-red & 0.6125 & 0.5200 & 0.5525 & 0.4800 & 0.9500 & 0.6275 & 2.6350 & 0.9950 \\
			SWD & 0.5240 & 0.4960 & 0.5400 & 0.4880 & 0.5640 & 0.5040 & 1.2040$\checkmark$& 0.8200$\checkmark$\\
			contact-lenses & 	0.5000 & 0.3333 & 0.5000$\checkmark$& 0.3333$\checkmark$ & 0.8333 & 0.8333 & 0.5000 & 0.3333 \\
			ESL & 0.4918 & 0.4344 & 0.5000 & 0.4426 & 0.5164 & 0.4590 & 1.8032  &0.7623$\checkmark$\\
			LEV & 0.5720 & 0.5120 & 0.5840 & 0.5120 & 0.700 & 0.5720  &  1.4120$\checkmark$& 0.8040 \\
			Automobile  & 0.9423$\checkmark$  & 0.5961$\checkmark$& 1.1346$\checkmark$ & 0.7692$\checkmark$ & 0.9808$\checkmark$& 0.7115$\checkmark$ & 2.8269 & 0.9808 \\
			Pasture & 1 & 0.6667 & 0.2222$\checkmark$& 0.2222$\checkmark$ & 0.6667 & 0.6667 & 1 & 0.6667  \\
			Squash-stored & 0.7692 & 0.6154 & 0.3846$\checkmark$ & 0.3077$\checkmark$& 0.5385 & 0.5385 &  0.7692 & 0.6154 \\
			\hline  
		\end{tabular}
	}
\end{table}

\begin{table}[!h]
	\caption{Performance of ocapis Instance selector} \label{tab:5}
	\noindent\resizebox{\linewidth}{!}{
		\begin{tabular}{ c c c c c c c c c c c l }
			\cline{2-9}
			&  \multicolumn{2}{c}{SVM}
			&  \multicolumn{2}{c}{POM} 
			&  \multicolumn{2}{c}{KDLOR}
			&  \multicolumn{2}{c}{WKNNOR}  \\
			\hline
			Dataset & MAE & MZE & MAE & MZE & MAE & MZE & MAE & MZE \\
			\hline	
			balance-scale &	0.1338 & 0.1274 & 0.1911 & 0.1401 & 0.4522 & 0.4331 & 0.4458 & 0.2675 \\
			winequality-red & 0.5150 & 0.4325 & 0.4400 & 0.4025 & 0.5075 & 0.4575$\checkmark$& 2.6350 & 0.9950 \\
			SWD & 0.4280$\checkmark$& 0.4120$\checkmark$ & 0.4760$\checkmark$& 0.4640 & 0.4920$\checkmark$& 0.4480$\checkmark$& 1.2800$\checkmark$& 0.8120$\checkmark$\\
			contact-lenses & 1 & 1 & 1$\checkmark$& 0.5000$\checkmark$& 0.5000 & 0.8333 & 0.5000 & 0.3333 \\
			toy & 1.1467 & 0.6533 & 1.1200  &  0.6400$\checkmark$& 0.5467 & 0.4267 & 1.8800$\checkmark$& 0.8800$\checkmark$\\
			ESL & 0.5328 & 0.5164 & 0.4262 & 0.3934 & 0.6475  & 0.5819 & 2.5246  & 0.9508 \\
			LEV & 0.4520  & 0.3960$\checkmark$ & 0.4400 & 0.3880 & 0.4800$\checkmark$& 0.4120$\checkmark$&  1.3960$\checkmark$& 0.7440$\checkmark$\\
			\hline  
	\end{tabular}}
\end{table}	

\begin{multicols}{2}
	usually the most expensive task in terms of computing time, \textbf{ocapis} performs well even when the number of features to select and the number of instances is high.
\section{Conclusions}
\label{sec:5}
Considering embedded order and monotonic restrictions present in ordinal and monotonic data is crucial when developing classification and preprocessing algorithms for data of that nature. \newline
In this paper we have presented the \textbf{ocapis} package for R. It was intended to provide efficient and scalable algorithms implemented in Scala for ordinal and monotonic data that are not yet available for researchers and practitioners of the R community. First, it includes two preprocessing techniques, an instance selector and a feature selector. Second, it includes four ordinal classification algorithms, one linear (Proportional Odds Models for Ordinal Regression) and three non-linear (Kernel Discriminant Learning for Ordinal Regression, Support Vector Machines with Ordered Partitions and Weighted k-Nearest-Neighbor for Ordinal Regression).\newline
As future work, we propose to keep maintaining and adding
algorithms for ordinal and monotonic data to our package, building a package to offer the vast majority of the major techniques proposed in the literature for ordinal regression and monotonic classification. Therefore, there are good perspectives to continue improving the software in the near future.

\begin{acknowledgements}
This work is supported by the Project BigDaP-TOOLS - Ayudas
Fundación BBVA a Equipos de Investigación Científica 2016.
\end{acknowledgements}

\section{Appendix. Installation Guide}
To install \textit{ocapis}, R language is needed (see the \href{https://www.r-project.org/}{R official site} for further  instructions on how to install it). Also, the required software includes a version of Python $\ge$2.7 (see \href{https://wiki.python.org/moin/BeginnersGuide/Download}{Python installation guide}), Scala $\ge$2.11 (see \href{https://www.scala-lang.org/download/}{Scala installation guide}) and libsvm-weights (\href{https://github.com/claesenm/EnsembleSVM/blob/master/libsvm-weights-3.17/README}{see libsvm-weights README}).
Once the requirements are satisfied, the latest developed version of \textit{ocapis} can be easily installed directly from Github through R with the \textit{devtools} package\cite{wickham2017devtools}:
\begin{minted}
[frame=lines,
framesep=1mm,
fontsize=\footnotesize,
bgcolor=bg,
linenos
]{r}
devtools::install_github("cristinahg/OCAPIS/OCAPIS")
\end{minted}

For further installation information, check the \href{https://cristinahg.github.io/OCAPIS/}{ocapis website}.
\bibliographystyle{spphys}       
\bibliography{references}   

%
%
\end{multicols}
\end{document}